\def\year{2022}\relax
\documentclass[letterpaper]{article} 
\usepackage{aaai22}  
\usepackage{times}  
\usepackage{helvet}  
\usepackage{courier}  
\usepackage[hyphens]{url}  
\usepackage{graphicx} 
\usepackage{natbib}  
\usepackage{caption} 
\DeclareCaptionStyle{ruled}{labelfont=normalfont,labelsep=colon,strut=off} 
\usepackage{algorithm}
\usepackage{algpseudocode}

\usepackage{newfloat}
\usepackage{listings}
\floatstyle{ruled}
\newfloat{listing}{tb}{lst}{}
\floatname{listing}{Listing}
\usepackage{subfigure}
\usepackage{booktabs}
\usepackage{multirow}
\usepackage{fixltx2e}
\usepackage{array}
\usepackage{amsmath}
\usepackage{amsfonts}
\usepackage{comment}
\usepackage{varwidth}

\newcommand{\stdv}[1]{\scriptsize$\pm$#1}

\usepackage{color} 

\title{Zero-Shot Out-of-Distribution Detection Based on the Pre-trained Model CLIP} 

\author {
    Sepideh Esmaeilpour\textsuperscript{\rm 1},
    Bing Liu\textsuperscript{\rm 1},
    Eric Robertson\textsuperscript{\rm 2},
    Lei Shu\textsuperscript{\rm 3}\footnote{Work was done prior to joining Amazon.}
}
\affiliations {
    \textsuperscript{\rm 1} University of Illinois at Chicago\\
    \textsuperscript{\rm 2} PAR Government\\
    \textsuperscript{\rm 3} Amazon AWS AI\\
    \{sesmae2, liub\}@uic.edu, eric\_robertson@partech.com, shulindt@gmail.com
}

\usepackage{bibentry}

\begin{document}

\maketitle

\begin{abstract}
        In an out-of-distribution (OOD) detection problem, samples of \textit{known classes} (also called \textit{in-distribution} \textit{classes}) are used to train a special classifier. In testing, the classifier can (1) classify the test samples of known classes to their respective classes and also (2) detect samples that do not belong to any of the known classes (i.e., they belong to some \textit{unknown} or OOD \textit{classes}).
        This paper studies the problem of \textit{zero-shot out-of-distribution} (OOD) \textit{detection}, which still performs the same two tasks in testing but has no training except using the given known class names.
        This paper proposes a novel and yet simple method (called ZOC) to solve the problem. ZOC builds on top of the recent advances in zero-shot classification through multi-modal representation learning. It first extends the pre-trained language-vision model CLIP by training a text-based image description generator on top of CLIP. 
        In testing, it uses the extended model to generate candidate unknown class names for each test sample and computes a confidence score based on both the known class names and candidate unknown class names for zero-shot OOD detection.
        Experimental results on 5 benchmark datasets for OOD detection demonstrate that ZOC outperforms the baselines by a large margin.
\end{abstract}

	\section{Introduction}\label{intro}
	The primary assumption in conventional supervised learning is that the samples encountered at the test time are from the same classes (called \textit{known} or \textit{seen classes}) that the model has observed and learned during training. However, this assumption, called the \textit{closed-world assumption}~\cite{fei2016breaking}, is often violated when a machine learning model is deployed in the real world; i.e., in addition to the seen classes, samples from unseen classes may appear at test. The seen class samples are referred to as the \textit{in-distribution samples} while unseen class samples are called \textit{out-of-distribution (OOD) samples}. It is crucial for an intelligent ML model to detect OOD samples specially in safety-critical applications such as autonomous driving or healthcare since detecting OOD samples as in-distribution ones in such applications can have catastrophic consequences.
	
	There are different directions in the literature tackling the OOD detection problem. The earlier methods are mainly based on SVMs~\cite{scheirer2012toward,scheirer2014probability,fei2016breaking}. Recent methods are mainly based on deep learning~\cite{shu2017doc,shu2018unseen,xu2019open,shu2021odist,liang2017enhancing,perera2020generative,miller2021class} and try to solve the problem from different perspectives. Some discriminative models perform the detection by calibrating the confidence of the \textit{closed-world classifier} built using seen or in-distribution classes~\cite{bendale2016towards}. \citet{liang2017enhancing} proposed to use temperature scaling on the softmax score and perform post-processing on the test data to detect the OOD data.~\citet{lee2017training} proposed a special training method for building closed-world classifiers that can also detect OOD samples at inference~\cite{lee2017training}. Some generative models synthesize samples to represent possible unseen classes~\cite{neal2018open}. These samples are then used to learn a $K+1$ classifier where the space of unseen is assumed to be enclosed in the extra class. Other generative methods also exist~\cite{andrews2016detecting,chen2017outlier}, which detect OOD samples based on the reconstruction error of the trained generative model for unseen samples.~\citet{perera2020generative} is a recent hybrid model based on generative-discriminative features.

	Regardless of the approaches, the results on OOD detection benchmarks indicate that the OOD detection performance is directly affected by the accuracy of the closed-world classifier. {Particularly, when the closed-world classifier does not use pre-trained models, it is essential to train an accurate classifier from scratch for a descent OOD detection performance. None of the aforementioned techniques use pre-trained models as the backbone of their closed-world classifiers. In fact, most of them essentially try to bound the hidden space representing the in-distribution classes. Then, the outer space can be considered as the OOD space.} 
	
	This paper defines the \textit{zero-shot OOD detection} problem to take advantage of pre-trained models. Given a set of seen class labels/names, $\mathcal{Y}_s$, the goal of zero-shot OOD detection is to 1) classify each seen class test sample to one of the seen classes and 2) detect samples that do not belong to any of the seen classes. These are done based on only the names of the seen classes in $\mathcal{Y}_s$. There is no given training data of the seen classes and thus no closed-world classifier is built. 

    CLIP~\cite{radford2021learning} is a recently proposed pre-trained language-vision model from OpenAI for zero-shot (closed-world) image classification. It is trained by directly using the raw text for learning visual representations. CLIP is a multi-modal (image and text) transformer model which is trained by contrastive learning on a large set of 400 million image and caption pairs collected from the Internet. The rich feature space shared by both image and text data enables zero-shot transfer to a range of down-stream tasks including image classification. CLIP model has an image encoder and a text encoder. Its zero-shot classification is done by matching the features from the image encoder to a set of text features from the text encoder. The text with the highest similarity score to the image is its predicted label. 
    
    Although using CLIP eliminates the need for training a closed-world classifier, it does not possess the OOD detection functionality in its original form. That is, it will match any given image to one of the given seen class labels. Therefore, to function in an OOD setting, we need to present another set of candidate labels in addition to the seen class labels/names. The proposed method, called ZOC (\textit{Zero-shot OOD detection based on CLIP}), does not need this set of candidate labels to represent possible OOD labels as ZOC can dynamically generate candidate OOD  labels for inference. ZOC works based on comparing the similarity of the semantic meaning of the given image to seen labels vs its similarity to some generated candidate labels. For this to work, we need a text generator to generate candidate labels, which does not exist in CLIP. To the best of our knowledge, existing OOD detection baselines either 1) need to train a closed-world classifier on seen classes using their labeled training data or 2) have prior knowledge about unseen classes for detection. ZOC requires neither of the two and therefore it is the first work performing zero-shot OOD detection. 
    In this work, we propose to:

    \begin{itemize}
    \item 
    Extend the CLIP model by training a textual description generator on top of CLIP's image encoder.
    \item Use the output of this generator as unseen candidate labels for a given test image.
    \item Define an OOD confidence score based on the similarity of the input test image to the union of the $seen$ labels and the $generated$ labels.
\end{itemize}
    Our experimental results show that this simple method outperforms many state-of-the-art fully supervised OOD detection baselines trained using benchmark datasets. {In addition to the supervised baselines, ZOC also outperforms the baselines that use the same pre-trained backbone model as ZOC.}

	\section{Related Work}
	\subsection{General Out-of-Distribution Detection}
    The terms Out-Of-Distribution (OOD) detection, Open Set Detection and Open World Classification are interchangeably used in the literature. In most papers about open set detection or open world classification, seen and unseen classes in evaluation are often two splits of the same dataset~\cite{fei2016breaking,shu2017doc,bendale2016towards,oza2019c2ae,perera2020generative,miller2021class,pernici2018memory,xu2019open}. For OOD detection, all seen classes (e.g., images of hand-written digits) are regarded as a single or multiple in-distribution classes, while the OOD data to be detected are from a different dataset (e.g., images of animals)~\cite{hendrycks2016baseline,liang2017enhancing,lee2018simple}. That is, the OOD classes are often visually completely dissimilar to in-distribution classes. 
    However, there is no fundamental difference between OOD detection and open set detection or open world classification.  This paper refers all of them as OOD detection.  
    
   We note that some OOD detection techniques are based on the idea of \emph{outlier exposure}. These methods either assume the direct access to a small subset of the actual test OOD data at training~\cite{liang2017enhancing} or rely on a large set of data points used as outliers at training~\cite{hendrycks2018deep}. However, most OOD detection methods, including ours, do not see any samples from unseen OOD classes before deployment. Recently, some authors made a distinction between hard and easy OOD detection problems~\cite{winkens2020contrastive}. That is, detecting OOD CIFAR100 from in-distribution CIFAR10 is considered as a \emph{near}-OOD (hard) problem as the two datasets contain visually similar categories. Likewise, detecting OOD CIFAR10 from in-distribution SVHN (photographed digits) is considered as a \emph{far}-OOD (easy) problem because their categories are visually and semantically very different. Despite this distinction, using a validation set from the OOD data to tune the model parameters, is a common practice in many OOD detection approaches. In this paper, we solve the near-OOD (or hard) problem in the zero-shot setting without using any validation OOD data. 
    
     \subsection{Transformer Model for OOD Detection}

     The success of pre-trained transformer models~\cite{vaswani2017attention, devlin2018bert} in the natural language domain has motivated researchers to analyze their performance for out-of-distribution or out-of-scope detection in real world applications.
     \citet{hendrycks2020pretrained} studies the OOD generalization and OOD detection performance of BERT for a range of NLP tasks. Their evaluation acknowledges that a pre-trained transformer improves OOD detection upon conventional models which are merely as good as a random detector for OOD detection. 

    The vision transformer model (ViT)~\cite{dosovitskiy2020image} works in a similar way to a language transformer, i.e., it divides an image to consecutive patches and then uses a regular transformer encoder to process these flattened patches as a sequence. ViT achieves on par or better performance than CNN-based methods like ResNets. A recent study~\cite{fort2021exploring} analyzed the reliability of OOD detection in ViT models. The authors show that ViT pre-trained models fine-tuned on an in-distribution dataset significantly improve near OOD detection tasks. In addition, this work is the most related work to ours in the sense that it performs zero-shot OOD detection through CLIP. However,~\citet{fort2021exploring} assumed that a set of unseen labels are given as some weak information about OOD data which is not practical in real world scenarios.

	\section{Method}
	\begin{figure*}[t]
    \centering
    \includegraphics[width=\textwidth]{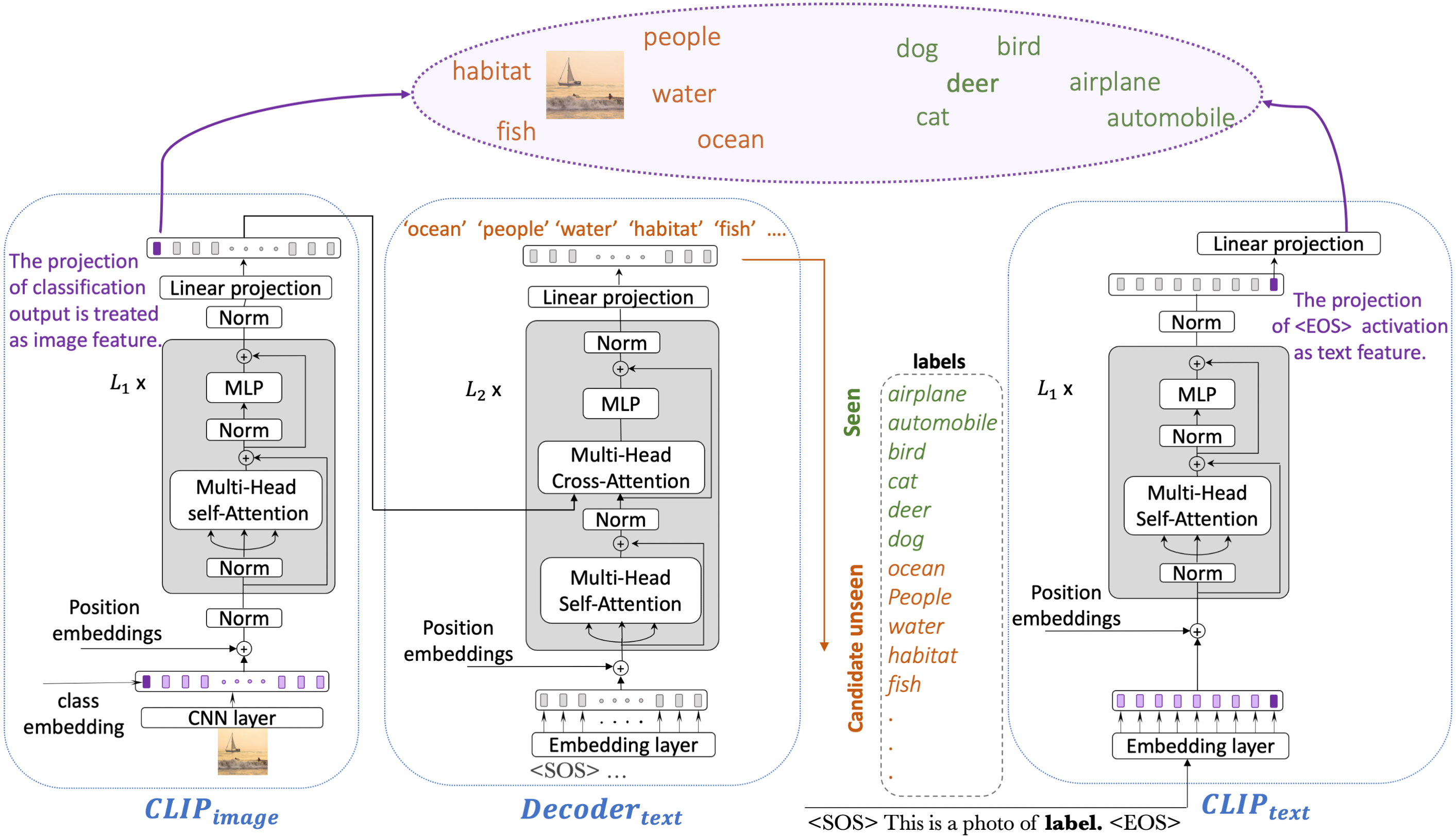}   
    \caption{The diagram illustrates the inference steps of ZOC for a sample from an unseen class `boat'. The available seen class labels (shown in green) are $\mathcal{Y}_s$=\{`airplane', `automobile', `bird', `cat', `deer', `dog'\}. In the first step, the image is encoded through CLIP\textsubscript{image} and then image description is generated in the output of Decoder\textsubscript{text}. The description is in fact a set of candidate unseen labels $\mathcal{Y}_u$ (shown in orange). In the second step, $\mathcal{Y}_s \cup \mathcal{Y}_u$ are encoded through CLIP\textsubscript{text} on the right.
    The purple ellipsoid shows CLIP's feature space where the relevant labels are aligned with the image. CLIP quantifies the alignment by calculating the cosine similarity of each encoded label to the encoded image. Then $S(x)$ is obtained according to \ref{conf_score}. The score is high for this image as it is more similar to the set of $\mathcal{Y}_u$ than $\mathcal{Y}_s$. The inference relies on CLIP pre-trained encoders as well as $\mathcal{Y}_u$ generated by Decoder\textsubscript{text} (Best viewed in color). 
    } \label{diagram}
\end{figure*}
    
    We propose to solve the zero-shot OOD detection problem by extending zero-shot CLIP~\cite{radford2021learning}, which is a closed-world zero-shot classification method, to work in the OOD setting. As mentioned in the introduction, the zero-shot CLIP model is not equipped with a specialized technique for OOD detection. Although for any given closed-world classifier, maximum softmax probability (MSP)~\cite{hendrycks2016baseline} is commonly used as a baseline score for OOD detection, we show in our experiments that our proposed method ZOC can significantly improve the detection performance. 
    ZOC detects an OOD test sample by comparing {the encoded image sample to two sets of encoded label names.} The first set is the set of seen labels, and the second set is the set of unseen labels which are unknown. ZOC trains a text description generator to obtain the second set. In the following, we briefly explain CLIP's matching algorithm for closed-world zero-shot classification and discuss its shortcomings for OOD detection. 
    
    For zero-shot closed-world classification in CLIP, we are only given a set of textual words as class labels $\mathcal{Y}_s=\{y_1, y_2, ..., y_n\}$. 
    For a test image, the multi-modal CLIP calculates the cosine similarity of the encoded image to each encoded textual description in the form of ``\emph{\{This is a photo of a $<y_i>$\}},'' e.g., ``\emph{This is a photo of a \underline{dog},'' ``This is a photo of a \underline{cat},'' etc.} Taking the softmax over all the $n$ similarity scores gives a categorical probability distribution that determines the label for the image. It is easily seen that any given image can be matched to one of the \emph{given} (possibly irrelevant) labels based on the maximum softmax score. As we can see, this method does not deal with zero-shot OOD detection. To do so, we propose to present CLIP with another set of possible labels $\mathcal{Y}_u$ for each test image sample for zero-shot matching. For this, we need a text-based image description generator. We train such a generator and use it to extract $\mathcal{Y}_u$ from a given test image. The next question is how the second set $\mathcal{Y}_u$ can assist in detecting an OOD sample. We will show later how the seen (known) labels together with the dynamic set $\mathcal{Y}_u$ can be used to define a confidence score per test image.
    
    Since CLIP does not have a text generator that can generate $\mathcal{Y}_u$ for a given image, we propose to train one on top of CLIP's image encoder using a large image captioning dataset. We explain the training of the generator next. We also call the \textit{text generator} the \textit{image description generator}.
\subsection{Training the Image Description Generator}
    Since our image description generator uses the output features from the CLIP image encoder for training, we briefly describe the CLIP image encoder here. CLIP uses ResNet-50~\cite{he2016deep} and the recently proposed vision transformer (ViT)~\cite{dosovitskiy2020image} as its image encoder backbone. We found that the ViT backbone is more compatible with the task of sequence generation from a given image since it processes the image as a sequence of tokens similar to the transformer model~\cite{vaswani2017attention}. The ViT encoder in CLIP is a hybrid ViT architecture which uses a convolutional layer in the beginning to extract image features. Then, $N$ feature maps are used as $N$ embedding vectors to represent the image as a sequence of embeddings. A classification embedding vector is concatenated to the image embeddings similar to the CLS token in BERT model~\cite{devlin2018bert}. Then,  positional embeddings are added and the output is passed to a transformer encoder~\cite{vaswani2017attention} with $L_1$ layers. The hidden state $z^{out}$ in the output is treated as the semantic representation of the whole image. 

    We train the text generator on a large image captioning data which is a set of image and caption pairs. Text generator, which is a decoder, attends to the encoder output feature $z^{out}$ in every layer of the decoder (see below). Please refer to Figure \ref{diagram} for architecture details. Text decoder consists of $L_2$ 
    stacked transformer layers. In each layer, the multi-head cross-attention sub-layer takes $z^{out}$ as key and value for the scaled dot product attention mechanism.
    The output from the final layer of the decoder is projected through a linear layer to the vocabulary space of the decoder.
    
    Assuming the text decoder is parameterized with $\theta$, the objective that we optimize is the  cross-entropy loss at each position $t$ in the sequence, conditioned on all previous positions:
    \begin{equation}
    L_{CE}(\theta)=-\sum_{t=1}^T log(p_{\theta}(y_t^*|y_{1:t-1}^*; z^{out}))
    \label{decoder_loss}
    \end{equation}
    This objective is optimized by forcing the predictions to stay close to the ground-truth sentence, which is the basic teacher forcing algorithm~\cite{williams1989learning}, i.e., the model simply conditions its next word prediction on previous ground-truth words (not previous \textit{predicted} words).

    As we will explain in the next section, the output description from the decoder will eventually be processed to be used at the second step of inference. Therefore, a generated description with relevant words to the image is sufficient for our purpose. We refer to the decoder as Decoder\textsubscript{text} in the rest of the paper. Decoder\textsubscript{text} outputs a textual description for a given image based on the hidden state of the CLIP image encoder which we refer to as CLIP\textsubscript{image}. In this regard, the image to sequence architecture is a full transformer model which has CLIP\textsubscript{image} on the encoder side and Decoder\textsubscript{text} on the decoder side (see Figure \ref{diagram}).

    \subsection{Inference in Testing}
        \begin{algorithm}[t]
	\caption{Zero-shot Open-Set Detection}\label{algo}
	\begin{algorithmic}[1]
		\Require set of seen labels $\mathcal{Y}_s$,  CLIP\textsubscript{text}, CLIP\textsubscript{image},  Decoder\textsubscript{text}, COCO dataset, set of test images $D^{test}$.
        \State \textbf{Training}
        \State initialize Decoder\textsubscript{text} model for sequence generation
        \State fine-tune Decoder\textsubscript{text} on COCO captioning dataset 
        \State \textbf{Inference}
        \For{$x^{test} \in D^{test}$}
            \State $\ \ \ \ $ labels=$\mathcal{Y}_s$
            \State $ \ \ \ \ $ description = Decoder\textsubscript{text}($x^{test}$).
            \State  $\ \ \ \ $ $\mathcal{Y}_u = topk(description)$
            \State $\ \ \ \ $ labels = $\mathcal{Y}_s \cup \mathcal{Y}_u$
            \State $\ \ \ \ $ logits $\gets$ $\emptyset$
             \For{label $\in$ labels}
                \State  desc = \textit{`This is a photo of a \underline{label}.'}
                \State sim = cosine(CLIP\textsubscript{image}($x_{test}$), CLIP\textsubscript{text}(desc))
                \State logits =logits $\cup$ sim 
            \EndFor
            \State  P = Softmax(logits)
            \State  $S(x^{test})=1-\sum_{y \in \mathcal{Y}_s}P(y|x)$ 
        \EndFor

	\end{algorithmic}
\end{algorithm}

    Decoder\textsubscript{text} is the central component for inference (testing) in our ZOC. The inference is performed in \textbf{lines 4-18} in Algorithm \ref{algo} which is composed of two steps. In the first step, Decoder\textsubscript{text} generates the image description for the given test image by attending to the image semantic representation in the output of CLIP\textsubscript{image}.
   {The generation follows the standard procedure of sequence to sequence models (predicting the next word based on the output of the model for the previous words until the maximum length is reached). ZOC needs to retrieve candidate unseen labels $\mathcal{Y}_u$ from the generated description.\footnote{{The generated description may contain stopwords, function words, etc. Since these words are present in every description, excluding/including them in $\mathcal{Y}_u$ does not affect the AUROC score which is calculated based on the ranking of confidence scores.}} Since $\mathcal{Y}_u$ is eventually used to define the confidence score for OOD detection, we would like the retrieved words to be diverse and relevant to the input image. i.e, diversity results in a more reliable confidence score for detection. However, canonical inference methods such as greedy generation, beam search, nucleus sampling~\cite{holtzman2019curious} or top-k sampling \cite{fan2018hierarchical} targets to generate the best 
    description rather than diverse  descriptions. Since we need a holistic description of the image in general, the best description does not suit our purpose as it is not diverse enough.
    Thus, we do not limit the set of candidate labels $\mathcal{Y}_u$ to be the same as the best generated description. 
    Instead, we form $\mathcal{Y}_u$ with some post-processing as follows: assuming the maximum generation length is $T$, at each position $p_i$ of $\{p_1, p_2, ..., p_T\}$, we pick the top $k$ words from the vocabulary with highest probabilities. The union of all these words is $\mathcal{Y}_u$ \textbf{(line 8 in Algorithm 1)}. We fix $k$ for all of our experiments. Then, we form the union of seen labels $\mathcal{Y}_s$ and candidate unseen labels $\mathcal{Y}_u$ \textbf{(line 9)}.}
    
    The second step follows the CLIP zero-shot classification technique based on zero-shot labels $\mathcal{Y}_s\cup \mathcal{Y}_u$. Each $y_i \in \mathcal{Y}_s \cup \mathcal{Y}_u$ is put in the template (i.e., “{\textit{This is a photo of a} $< y_i>$}) required by CLIP. The text and the image are encoded through CLIP\textsubscript{text} and CLIP\textsubscript{image} and the cosine similarity of the encoded image and encoded label (in template) is calculated \textbf{(lines 11-15).} The softmax of all calculated similarities gives a probability distribution over $\mathcal{Y}_s \cup \mathcal{Y}_u$ \textbf{(line 16).}
    We define the OOD confidence score \textbf{(line 17)} as follows:
    \begin{align}\label{conf_score}
            S(x) = 1-\sum_{y \in \mathcal{Y}_s}P(y|x)
    \end{align}
where $P(y|x)$ is the softmax probability for label $y$. Thus, $S(x)$ is the accumulative probability of labels $\mathcal{Y}_u$. Even though ZOC inference is done in two steps, the implementation and usage of our technique is straightforward as the second step is done by querying the CLIP encoders.

    {Figure \ref{diagram} is a graphical illustration of the inference procedure of ZOC. The used example describes how ZOC detects a sample as OOD. The input image is from class `boat' which is not among the seen labels and therefore it is an unseen class or OOD sample. It is interesting to note that the actual unseen label `boat' is not among the set of candidate unseen labels, and yet ZOC uses other candidate unseen labels to come to the correct conclusion.} 
    
	\section{Experiments}
	\subsection{Model Architecture and Training Details}
    Recall that ZOC consists of 3 modules. The two encoders CLIP\textsubscript{image} and CLIP\textsubscript{text} are pre-trained transformer models for image and text~\cite{radford2021learning}, respectively. We do not change or fine-tune the encoders. CLIP\textsubscript{text} is a base transformer model with 12 stacked layers and hidden size of 768. The final linear projection layer outputs a representation of size 512. CLIP\textsubscript{image} is a hybrid ViT-base model using a convolutional layer in the beginning for feature extraction. The images are center-cropped and resized to size 224*224. A total of 7*7=49 embedding vectors with hidden size of 768 are generated from a given image. The transformer encoder in ViT also has 12 stacked layers. The output hidden state is projected from 768 to 512 dimensions to have the same size as CLIP\textsubscript{text}. For the proposed Decoder\textsubscript{text}, we choose the BERT large model from huggingface~\cite{wolf-etal-2020-transformers} with 24 layers and hidden size of 1024. We train Decoder\textsubscript{text} using Adam optimizer~\cite{kingma2017adam} with a constant learning rate of $10^{-5}$ for 25 epochs. Batch size is 128. The training data for fine-tuning is the training split of MS-COCO (2017 release)~\cite{lin2014microsoft}\footnote{https://cocodataset.org} which is a commonly used dataset for image captioning. We used MS-COCO validation dataset to choose the $k$ value. We empirically found that the meaningful candidate unseen labels are present at top $35$ level of the annotations. We used the basic teacher forcing method to train Decoder\textsubscript{text} as it is sufficient for our purpose. There are other principled sampling approaches such as scheduled sampling~\cite{bengio2015scheduled}, professor forcing~\cite{lamb2016professor}, and self-critical training~\cite{rennie2017self} for training a sequence generation model. These approaches try to alleviate the exposure bias in testing, which is not our concern.
    
\subsection{Datasets}
 We evaluate the performance of our proposed method ZOC on splits of CIFAR10, CIFAR100, CIFAR+10, CIFAR+50, and TinyImagenet. The difficulty level of an OOD detection task is commonly measured by the \textit{openness metric} defined in~\cite{scheirer2012toward}. A task is harder when more unseen classes are presented to the model at the test time. Openness is defined as follows 
\begin{equation}
Openness = ( 1-\sqrt{\frac{2*N_{train}}{N_{test}+N_{target}}})*100
\end{equation}
where $N_{train}$ is the number of seen classes, $N_{target}$ is the number of seen classes at testing and $N_{test}$ is the total number of seen and unseen classes at test.
For \emph{CIFAR10}~\cite{Krizhevsky2009learning}\footnote{https://www.cs.toronto.edu/~kriz/cifar.html} 6 classes are used as in-distribution (or seen) classes. The 4 remaining classes are used as OOD (unseen) classes. The reported score is averaged over 5 splits (Openness = $13.39\%$).
For \emph{CIFAR+10}~\cite{Krizhevsky2009learning}\footnote{https://www.cs.toronto.edu/~kriz/cifar.html} 4 non-animal classes of CIFAR10 are used as in-distribution (or seen) classes. 10 animal classes are chosen from CIFAR100 as the OOD (unseen) classes. The reported score is averaged over 5 splits (Openness = $33.33$).
For \emph{CIFAR+50}~\cite{Krizhevsky2009learning}\footnote{https://www.cs.toronto.edu/~kriz/cifar.html} 4 non-animal classes from CIFAR10 are in-distribution (or seen). All 50 animal classes from CIFAR100 are used as the OOD classes (Openness = $62.86\%$).
For \emph{TinyImagenet.}~\cite{le2015tiny}\footnote{http://cs231n.stanford.edu/tiny-imagenet-200.zip} 20 classes are used as the in-distribution (or seen) classes. The remaining 180 classes are used as OOD (unseen) classes. The reported score is averaged over 5 splits (Openness = $57.35\%$).
For \emph{CIFAR100}~\cite{Krizhevsky2009learning} \footnote{https://www.cs.toronto.edu/~kriz/cifar.html} 20 classes are used as in-distribution (or seen). The 80 remaining classes are used as OOD classes. The reported score is averaged over 5 splits. In each split, 20 consecutive classes are used as seen and the rest of classes used as unseen (Openness = $42.26\%$). The class splits that we have used are publicly available in the github repository of~\cite{miller2021class} \footnote{https://github.com/dimitymiller/cac-openset} for all datasets except for CIFAR100. We generated the splits for CIFAR100 as explained above.

\subsection{Baselines}
\begin{table*}[ht!]
    \centering 
    \resizebox{0.9\linewidth}{!}{
    \begin{tabular}{lcccccc}
    \toprule
    &CIFAR10&CIFAR100&CIFAR+10&CIFAR+50&TinyImageNet&Average\\
    \midrule
      \multicolumn{2}{l}{\textbf{Original baselines$\to$}}&\multicolumn{5}{c}{}\\
     \hline
         OpenMax~\cite{bendale2016towards} & 69.5\stdv4.4& NR &81.7\stdv NR& 79.6\stdv NR & 57.6\stdv NR& 75.6\\
         DOC \cite{shu2017doc} & 66.5\stdv6.0 & 50.1\stdv0.6 & 46.1\stdv1.7 & 53.6\stdv0.0 & 50.2\stdv0.5&58.2 \\
         G-OpenMax~\cite{ge2017generative} & 67.5\stdv4.4 &NR & 82.7\stdv NR & 81.9 \stdv NR &  58.0 \stdv NR&75.9 \\
        OSRCI~\cite{neal2018open} & 69.9\stdv3.8 &NR & 83.8\stdv NR & 82.7\stdv0.0 & 58.6\stdv NR&77.2 \\
        C2AE~\cite{oza2019c2ae}
        & 71.1\stdv0.8 & NR & 81.0\stdv0.5 & 80.3\stdv0.0 & 58.1\stdv1.9&75.9 \\
        GFROR~\cite{perera2020generative} & 80.7\stdv3.0 &NR & 92.8\stdv0.2 & 92.6\stdv0.0 & 60.8\stdv1.7&84.0 \\
       CSI~\cite{tack2020csi} & 87.0\stdv4.0 & 80.4\stdv1.0 & 94.0\stdv1.5 &97.0\stdv0.0 & 76.9\stdv1.2& 87.0 \\    
        CAC~\cite{miller2021class} & 80.1\stdv3.0 & 76.1\stdv 0.7& 87.7\stdv1.2 & 87.0\stdv0.0 & 76.0\stdv1.5&84.9 \\
     \hline
     \multicolumn{3}{l}{\textbf{Baselines with CLIP backbone/initialization $\to$}}&\multicolumn{4}{c}{}\\
     \hline
        CLIP+CAC~\cite{miller2021class} & 89.3\stdv2.0 & \textbf{83.5\stdv 1.2}& 96.5\stdv0.5 & 95.8\stdv0.0 & 84.6\stdv1.7&89.9 \\
        CLIP+G-ODIN~\cite{hsu2020generalized} & 63.4\stdv3.5 & 79.9\stdv2.3 &45.8\stdv1.9 & 92.4\stdv0.0  &67.0\stdv7.1&69.8\\
        CLIP+MSP~\cite{hendrycks2016baseline}& 88.0\stdv3.3 & 78.1\stdv3.1 &94.9\stdv0.8 & 95.0\stdv0.0 & 80.4\stdv2.5&87.3\\
    \hline
        \emph{ZOC (ours})& \textbf{93.0\stdv1.7} & 82.1\stdv2.1 &\textbf{97.8\stdv0.6} &\textbf{97.6\stdv0.0} & \textbf{84.6\stdv1.0}&\textbf{91.0}\\
    \bottomrule
    \end{tabular}
    }
     \caption{
     OOD detection performance in AUROC. The first 8 rows give the results of the original versions of the supervised baselines which train a separate classifier for each set of in-distribution classes. We have also combined CLIP with three systems, denoted by CLIP+X. We initialized the weights of CAC and G-ODIN and then fine-tuned their closed-world classifiers. For MSP, we simply used the pre-trained CLIP encoder to generate softmax scores. Each result in the table is the average of 5 splits of each dataset ($\pm$ standard deviation).
     }\label{AUC performance} 
\end{table*}

We compare our method with 11 OOD detection baselines. Each baseline either requires to train a closed-world classifier or works based on a pre-trained model as its backbone. In both cases, labeled training data is required. We are not aware of any existing zero-shot OOD detection model except~\cite{fort2021exploring} which requires unseen class labels to be given for detection (the paper's main focus is not zero-shot OOD detection). Therefore, it is unsuitable for our OOD detection setting, and thus is not included as a baseline. 

\emph{DOC}~\cite{shu2017doc} is an early method originally proposed for OOD detection (or recognition) of text data. It uses one-vs-rest sigmoid function in the output layer. It compares the maximum score over sigmoid outputs to a predefined threshold to reject or accept a test sample. \emph{OpenMax}~\cite{bendale2016towards} is an early technique for OOD image recognition. It does calibration on the penultimate layer of the network to bound the open space risk.
\emph{G-OpenMax} and \emph{OSRCI} \cite{ge2017generative,neal2018open} are both generative models that use a set of generated samples to learn an extra class. So, the model is a $K+1$ class classifier of seen and pseudo-unseen. 
\emph{C2AE} \cite{oza2019c2ae} is a class-conditioned generative method that uses the reconstruction error of test samples as the detection score. 
\emph{CAC}~\cite{miller2021class} is a latest method that uses anchored class centers in the logit space to encourage forming of dense clusters around each known/seen class. Detection is done based on the distance of the test sample to the anchored seen class centers in the logit space.
\emph{GFROR} \cite{perera2020generative} combines the advantage of generative models with the recent advances in self-supervision learning methods.
\emph{G-ODIN} \cite{hsu2020generalized} {is a recent method that uses a decomposed confidence score on top of its feature extractor for OOD detection. The hyperparameters of the algorithm are tuned only on closed-world classes.} 
\emph{CSI}~\cite{tack2020csi} 
combines contrastive learning, self-supervised learning, and various data augmentation techniques to train its model. It is a latest strong baseline.
{\emph{MSP} \cite{hendrycks2016baseline} uses maximum softmax probability as the natural OOD detection score which can be used on top of any closed-world classifier. Ideally, MSP is high on in-distribution (or closed-world) classes and low for OOD classes. 
}
{The results for OpenMax, G-OpenMax, C2AE and CAC are taken from \cite{miller2021class}. We adapted DOC's code for images to generate its results. We ran the official code of CAC for CIFAR100. All these baselines use a CNN encoder architecture introduced in \cite{neal2018open}. We ran CSI's official code to produce its results.
We also tried to combine CLIP and some baselines. Since the code of G-ODIN is not released, we implemented it following its algorithm and hyper-parameters. For fair comparison with ZOC, we used the image encoder of CLIP as G-ODIN's backbone (denoted by CLIP+G-ODIN). We further created a version of CAC using CLIP to initialize its weights and then fine-tuned its classifier (denoted by CLIP+CAC). CAC was chosen as it is compatible with CLIP and is on average the second best performing baseline after CSI. For MSP, which can be used on top of any trained classifier, we used CLIP zero-shot classification pipeline to generate the results.
For CSI, since it learns a specific feature extractor based on 0, 90, 180, 270 degree rotations of every sample, it is incompatible with the pre-trained model CLIP.
} 
\subsection{Results and Discussion}
\begin{figure*}[t]
\centering
\subfigure[Error study of generated labels]{%
{\includegraphics[width=0.52\linewidth, height=0.4\linewidth]{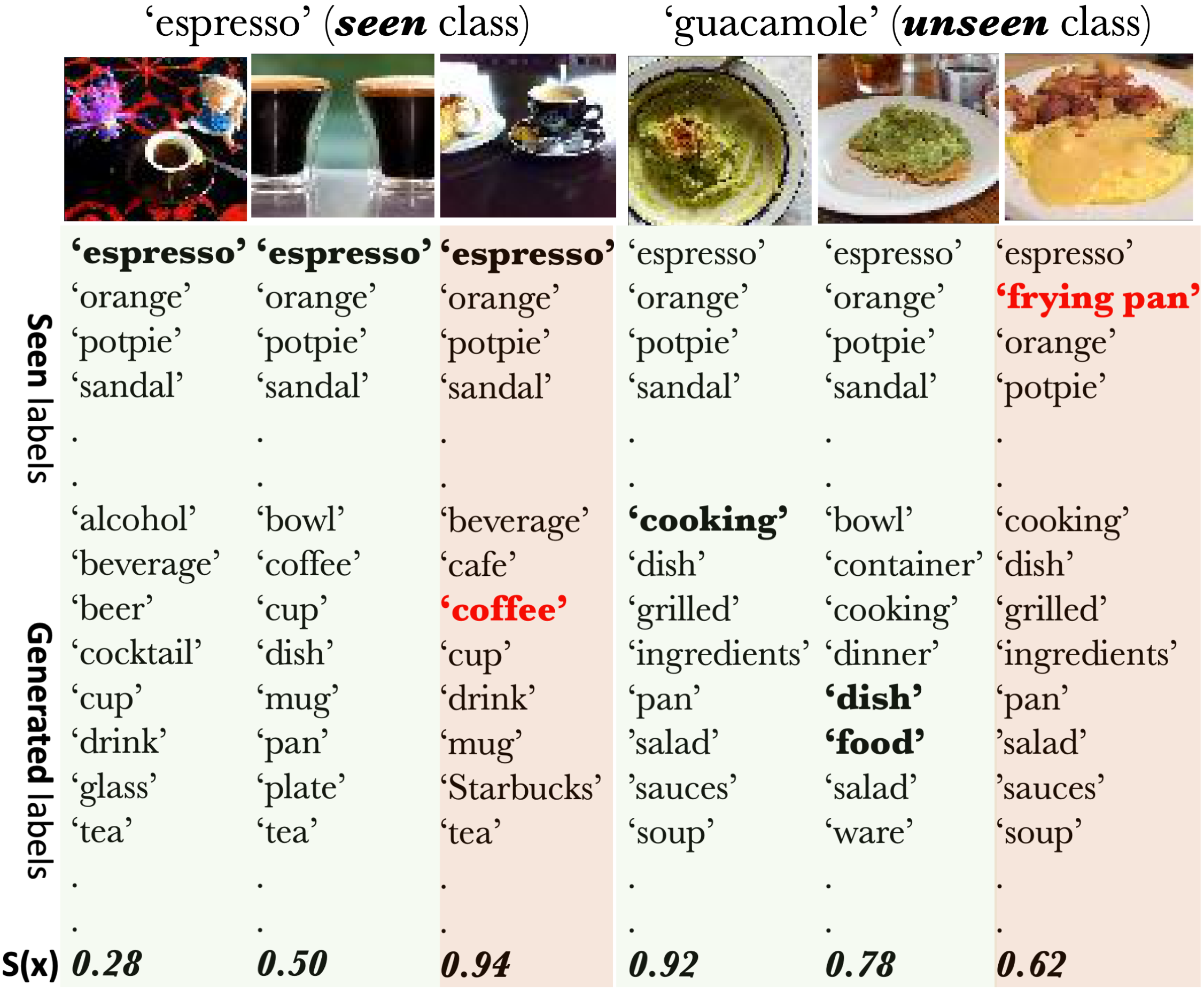}%
\label{errorstudy}%
}}\qquad
\subfigure[Histograms of confidence scores]{%
\includegraphics[width=0.43\linewidth, height=0.4\linewidth]{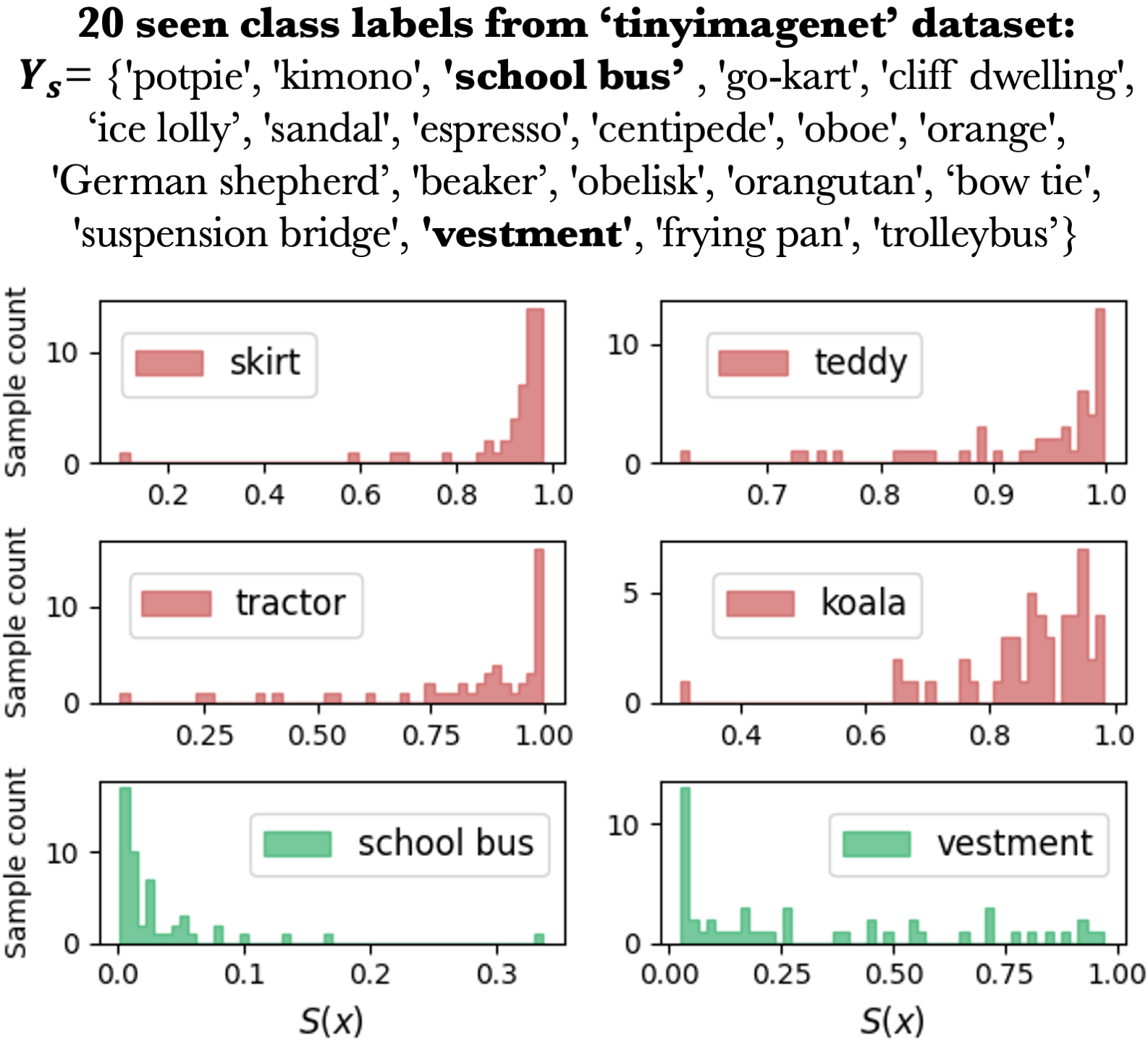}%
\label{histogram}%
}
\caption{(a) A summary of the generated labels for a seen class `espresso' and an unseen class `guacamole' from the TinyImagenet dataset are shown. The generated labels are ranked based on their contribution to $S(x)$. The labels with ($P(y|x)>0.1$) are in boldface. For class `espresso', we expect the model to output a relatively low $S(x)$ as the actual label is present among seen labels (first two images). The third image is an error case. The set of generated labels and the label `coffee' produce a high $S(x)$. For the unseen class `guacamole', $S(x)$ is high for the first two images as expected since ZOC correctly associates the generated labels with the images. The third image is again an error case when a seen label `frying pan' contributes to $S(x)$ more than the generated unseen labels. (Best viewed in color) (b) 20 seen labels form tinyimagenet are listed at the top. 4 classes `skirt', `teddy', `tractor' and `koala' are a subset of unseen classes. Each subplot shows the histogram of the confidence score $S(x)$. For instance, in the histogram for unseen class `skirt', we can clearly see that more than 40 samples have confidence scores between 0.8 and 1 which is desirable for good detection performance. S(x) tends to have a higher variance for the other 3 unseen class plots. It is interesting to note that for the samples from class `tractor', the confidence score is relatively low because it is confused with semantically similar seen labels `school bus' and `go-kart'. Similarly, low $S(x)$ for `koala' mostly happen when the model associates the image with seen labels `orangutan' and `German shepherd'. The confidence score for two seen classes `school bus' and `vestment' is distributed in lower ranges as expected. (Best viewed in color)}
\end{figure*}

The experimental results\footnote{ https://github.com/sesmae/ZOC} are summarized in Table~\ref{AUC performance}.
We use AUROC (Area Under the ROC curve) as the evaluation measure as it is the most commonly used measure for OOD detection. 
ZOC outperforms all baselines by a large margin. A significant difference between ZOC and the baselines is that ZOC inference is based on dynamically generated candidate unseen labels for each sample, which gives ZOC a better detection capability.

Since ZOC uses CLIP's pre-trained encoder for inference, one might attribute the performance gain to the rich feature space of CLIP. To investigate the gain, we set up an experiment with MSP~\cite{hendrycks2016baseline} for CLIP. As mentioned earlier, MSP uses the maximum softmax score of zero-shot CLIP as the OOD confidence score. This experimental setup assesses CLIP's inherent ability for OOD detection compared to ZOC's inference technique. ZOC's consistent performance gain over MSP on all datasets (Table \ref{AUC performance}) confirms that the proposed confidence score based on dynamic generation of unseen labels is better than MSP which uses identical inference procedure for all samples.

Note that we do not have an ablation study on ZOC as no part of the algorithm can be dropped for it to function. 
\subsubsection{Case study and error analysis.}
Figure \ref{errorstudy} is a case study illustrating the actual seen and candidate unseen labels (generated for each sample) used in calculating the confidence score $S(x)$.  We picked one seen and one unseen class for comparison. Note that the actual label for a given image might/might not be among the generated labels. Particularly, this can happen when the unseen label is fine-grained and not present in the training corpus of MS-COCO.  As a result, for an image with label `espresso', the decoder generates relevant words such as `coffee' rather than the label `espresso' itself. Nevertheless, ZOC comes to the correct conclusion based on accumulative confidence score $S(x)$. Figure \ref{histogram} illustrates the statistics of the calculated confidence score for 4 unseen and 2 seen classes from tinyimagenet. 
We plan to use a larger corpus to train the image description generator in the future. In addition, since ZOC compares standalone candidate labels to the image, it does not account for relations between the unseen labels. Such relations might be an important tool for detecting more sophisticated OOD  samples. We will address this limitation in our future work. 
\section{Conclusion}
    {In this paper, we introduced the new task of \textit{zero-shot OOD detection} based on the recent advances in zero-shot closed-world classification using the pre-trained model CLIP~\cite{radford2021learning}. Since it is a zero-shot problem,  1) no concrete samples are given for training except the known or seen class label names, and 2) samples from unseen OOD classes may appear at the test time. 
    To solve the problem, we extended the CLIP model so that it can dynamically generate candidate unseen labels for each test image, and also defined a novel confidence score calculated based on the similarity of the test image to seen and generated candidate unseen labels in the feature space. Experimental results confirmed that the proposed system ZOC is superior to the traditional supervised models. In addition, it also outperforms the baselines which use pre-trained CLIP backbone as their encoders.}  

\section*{Acknowledgments}
The work of Sepideh Esmaeilpour and Bing Liu was supported in part by a DARPA Contract HR001120C0023, two National Science Foundation (NSF) grants (IIS-1910424 and IIS-1838770), and a Northrop Grumman research gift. The work of Eric Robertson was supported in part by the DARPA Contract HR001120C0023.

\bibliography{AAAI22_ZeroShot_OOD.bib}

\end{document}